\PassOptionsToPackage{table}{xcolor}
\documentclass[]{fairmeta}

\title{\vspace{-1pt}BlockVLA: Accelerating Autoregressive VLA via Block Diffusion Finetuning}

\author[*]{Ruiheng Wang}
\author[*]{Shuanghao Bai}
\author{Haoran Zhang}
\author{Badong Chen}
\author[\dagger]{Xiangyu Xu}

\affiliation{Xi'an Jiaotong University}

\contribution[*]{Equal Contribution}
\contribution[\dagger]{Corresponding author}

\abstract{
While autoregressive (AR) Vision-Language-Action (VLA) models have demonstrated formidable reasoning capabilities in robotic tasks, their sequential decoding process often incurs high inference latency and may amplify error accumulation during long-horizon execution. 
Discrete Diffusion Language Models (dLLMs) provide a promising alternative through parallel token refinement, but their practical deployment in robotics remains limited by repeated denoising function evaluations (NFEs) and the difficulty of directly applying standard KV caching to bidirectional iterative decoding.
To bridge these paradigms, we propose BlockVLA, a framework that adapts pretrained AR backbones into an efficient discrete diffusion policy through a block diffusion paradigm. BlockVLA maintains autoregressive dependencies at the block level while enabling parallel denoising within each block, thereby combining global causal coherence with local parallel generation. This design enables prefix KV-cache reuse across completed blocks, reduces the effective cost of iterative denoising, and provides a smoother transition from AR pretraining to diffusion-based policy fine-tuning.
We conduct extensive evaluations on the LIBERO and SimplerEnv benchmarks. Experimental results demonstrate that our BlockVLA achieves a 3.3$\times$ inference acceleration over standard discrete diffusion baselines. Furthermore, our model exhibits superior training efficiency, with success rates converging substantially faster than baselines, a gain that is particularly pronounced in complex, long-horizon tasks, where BlockVLA achieves significant performance gains in the early stages of training. This work establishes Block Diffusion as a robust bridge between large-scale pretrained AR models and efficient, high-frequency real-time robotic control.
}

\usepackage{subcaption}
\usepackage{enumitem}
\usepackage{algorithm}
\usepackage{algorithmic}
\usepackage{multicol}
\usepackage{multirow}
\usepackage{wrapfig}
\usepackage{adjustbox}
\usepackage{minipage-marginpar}
\usepackage{amsmath}
\usepackage{amsthm}
\usepackage{array}
\usepackage{pgfplots}
\usepackage[utf8]{inputenc} 
\usepackage[T1]{fontenc}    
\usepackage{url}            
\usepackage{booktabs}       
\usepackage{pifont}
\usepackage{pbox}
\usepackage{xspace}
\usepackage{hyphenat}       
\usepackage{amsfonts}       
\usepackage{nicefrac}       
\usepackage{microtype}      
\usepackage{lipsum}
\usepackage{fancyhdr}       
\usepackage{graphicx}       
\graphicspath{{media/}}     

\theoremstyle{definition}

\theoremstyle{remark}

\definecolor{citecolor}{HTML}{0071BC}
\definecolor{linkcolor}{HTML}{ED1C24}
\definecolor{acceptcolor}{HTML}{74C219}
\definecolor{rejectcolor}{HTML}{DE1616}
\definecolor{qcolor}{HTML}{536872}
\definecolor{demphcolor}{RGB}{100,100,100}
\definecolor{brightlavender}{rgb}{0.75, 0.58, 0.89}
\definecolor{palered}{rgb}{1.00, 0.70, 0.70}
\definecolor{palegreen}{rgb}{0.73, 0.96, 0.67}
\definecolor{paleblue}{rgb}{0.69, 0.84, 1.00}
\definecolor{paleorange}{rgb}{1.00, 0.86, 0.73}
\definecolor{palepurple}{rgb}{0.92, 0.85, 1.00}
\definecolor{paleyellow}{rgb}{1.00, 1.00, 0.50}

\definecolor{commentgray}{RGB}{110,110,110}
\newcommand{\algcomment}[1]{{\color{commentgray}\# #1}}

\newlength\savewidth

\newcommand{\app}{\raise.17ex\hbox{$\scriptstyle\sim$}}

\definecolor{lightgray}{rgb}{0.95, 0.95, 0.95}

\definecolor{baselinecolor}{gray}{.9}

\setlist[enumerate]{itemsep=-0.5mm,partopsep=0pt}

\newcolumntype{x}[1]{>{\centering\arraybackslash}p{#1pt}}
\newcolumntype{y}[1]{>{\raggedright\arraybackslash}p{#1pt}}
\newcolumntype{z}[1]{>{\raggedleft\arraybackslash}p{#1pt}}
\pgfplotsset{compat=1.18}
\begin{document}

\maketitle

\section{Introduction}
	Autoregressive (AR) models have demonstrated exceptional performance in multimodal understanding via the next-token prediction paradigm. In the realm of discrete Vision-Language-Action Models (VLA) in robotics, frameworks (e.g. RT-2~\citep{vla:rt2}, OpenVLA~\citep{vla:openvla}, $\pi_0$-FAST ~\citep{vla:fast}) tokenize continuous actions into discrete tokens, enabling models to inherit the profound reasoning and grounding capabilities of web-scale pretrained Vision-Language Models (VLMs) ~\citep{vlm:prismatic,vlm:paligemma,vlm:qwen3,vlm:internvl} through large-scale robotic pretraining~\citep{dataset:oxe,dataset:bridge,dataset:droid,dataset:agibot} and subsequent supervised fine-tuning (SFT) on specific tasks~\citep{bai2025embodied}.
	However, the inherent sequential decoding of AR models imposes critical bottlenecks: high inference latency and compounding execution errors during long-horizon tasks, which frequently lead to catastrophic failure. To address these limitations, discrete Diffusion Language Models (dLLM)~\citep{ddvla:discrete,ddvla:lladavla,ddvla:dream,ddvla:udvla} have emerged as a promising alternative. As shown in ~\Cref{tab:model_comparison}, by leveraging discrete diffusion processes, dLLMs enable parallel decoding to accelerate inference while effectively mitigating cumulative errors.
    \begin{figure}[t]
        \centering
        \includegraphics[width=0.95\linewidth]{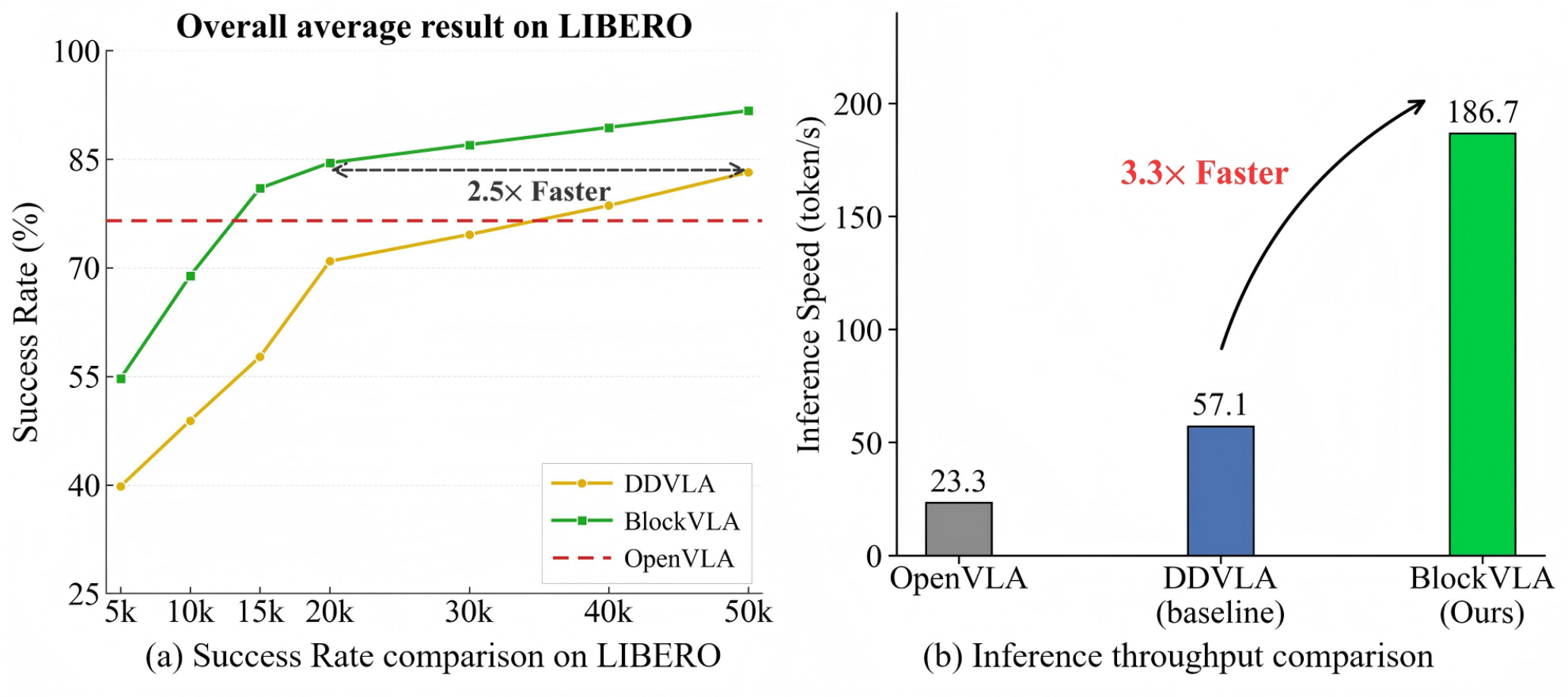}
        \caption{\textbf{Quantitative results on training efficiency and inference throughput.} (a) Average success rate on LIBERO: Our BlockVLA outperforms the baseline in overall average success rate across four task suites and exhibits superior training efficiency with significantly faster convergence. (b) Inference throughput comparison: Our BlockVLA achieves 3.3$\times$ acceleration than the DDVLA baseline on a single NVIDIA RTX 4090 GPU.}
        \label{fig:overall_results}
    \end{figure}
    While dLLMs, including dVLMs~\citep{dlm:d3pm,dlm:llada,dlm:llada2,dlm:dream,dvlm:lladav,dvlm:mmada}, promise to alleviate the sequential decoding bottleneck through parallel token refinement, their practical speedup in robotic applications remains constrained. 
    The iterative denoising process typically requires a large number of denoising function evaluations (NFEs), where each NFE denotes one model forward pass during refinement, to produce high-fidelity action sequences. 
    In addition, the bidirectional iterative decoding of conventional dLLMs limits the direct reuse of prefix states through standard KV caching~\citep{dlm:dllmcache,dlm:fastdllm}, often offsetting the throughput gains from parallelization.
    Beyond inference efficiency, adapting dLLMs to robotics also faces a model availability challenge. 
    Modern VLMs are still overwhelmingly dominated by AR backbones, while mature diffusion-native VLMs remain scarce. 
    Given the prohibitive cost of training dLLMs from scratch, adapting pretrained AR-based VLMs to the diffusion paradigm~\citep{ar2dlm:scaling,ar2dlm:diffusionvl,ar2dlm:sdar,ar2dlm:efficient} offers a practical way to inherit large-scale visual-linguistic knowledge. 
    Nevertheless, this adaptation is challenging because the causal factorization of AR models differs from the bidirectional iterative refinement used in diffusion models, creating optimization and deployment barriers for efficient robotic policy learning.

	\begin{table}[htbp]
		\centering
		\caption{Comparison of different modeling paradigms for VLA models.}
		\vskip 0.05in
		\label{tab:model_comparison}
			\begin{tabular}{lccc}
				\toprule
				\textbf{Feature} & \textbf{Autoregressive} & \textbf{Discrete Diffusion} & \textbf{Block Diffusion (Ours)} \\ \midrule
				Less Cumulative Error & \ding{55} & \ding{51} & \ding{51} \\
				KV-Caching           & \ding{51} & \ding{55} & \ding{51} \\
				Parallel Decoding    & \ding{55} & \ding{51} & \ding{51} \\ 
				Efficient Seq Scaling         & \ding{51} & \ding{55} & \ding{51} \\
				\midrule
				Inference speed      & Slow      & Fast      & \textbf{Faster} \\ \bottomrule
			\end{tabular}
	\end{table}
	
	To address these limitations, we propose \textbf{BlockVLA}, which adapts the block diffusion paradigm~\citep{bd:block,ar2dlm:efficient, bd:fastdlm2,bd:df} to robotic manipulation. Unlike standard dLLMs~\citep{bd:block,ar2dlm:sdar} that treat the entire sequence as homogeneous training targets, our framework accounts for the asymmetric nature of VLA training targets, where visual-linguistic inputs serve as persistent conditions and only actions are generated. We specialized the masking strategy to accommodate this asymmetry, ensuring that parallel action denoising remains strictly grounded in the multi-modal prefix to achieve superior training-inference consistency. Our method restructures action generation by maintaining an AR structure at the block level while performing parallel denoising within each block. This hybrid design preserves the causal logic and KV-Cache efficiency of AR models, ensuring temporal consistency. Simultaneously, intra-block parallel diffusion enables high-throughput action generation. Crucially, by retaining the global causal structure, Block Diffusion facilitates a smoother transition for adapting pretrained AR backbones than bidirectional approaches, effectively balancing reasoning stability with execution speed.

	We implement our framework using OpenVLA~\citep{vla:openvla} as the pretrained backbone and conduct extensive evaluations on the LIBERO~\citep{bench:libero} and SimplerEnv~\citep{bench:simplerenv} robotic benchmarks. As shown in ~\Cref{fig:overall_results}(a), our experimental results demonstrate that, with a limited training budget of 50k steps on only 2 GPUs, our BlockVLA achieves an average success rate of 91.7\% on LIBERO, significantly outperforming the baseline’s 83.2\%, with comparable gains on Object and Long suites. Notably, our Block Diffusion achieves an inference speedup of $3.3\times$ compared to standard discrete diffusion baseline, which is shown in ~\Cref{fig:overall_results}(b). Furthermore, our analysis of success rate curves across training steps reveals that Block Diffusion exhibits superior training efficiency, with success rates converging faster than the baseline. This advantage is particularly pronounced in complex, long-horizon tasks, where the synergy of causal block-wise structure and intra-block parallel generation allows the model to capture temporal dependencies more robustly and with substantially fewer training iterations.
	
	In summary, our primary contributions are as follows:
	\begin{itemize}[leftmargin=1.5em]
		\item We present \textbf{BlockVLA}, a novel framework that pioneers the adaptation of pretrained Autoregressive VLAs into a Discrete Diffusion paradigm through Block Diffusion. By reconciling block-level causality with intra-block parallelism, our method preserves KV-Cache while enabling high-throughput action generation.
		\item We provide a comprehensive investigation into the Block Diffusion training paradigm tailored for VLA's multi-modal asymmetry. By distinguishing visual-linguistic context as persistent conditions from action generation targets, we design a specialized masking strategy that ensures training-inference consistency. Our analysis of various attention patterns and masking behaviors offers actionable insights into bridging the gap between causal pretraining and bidirectional diffusion for action sequences.
		\item We conduct extensive experiments on the LIBERO and SimplerEnv benchmarks. Our results demonstrate that BlockVLA achieves approximately a $3.3\times$ inference acceleration compared to standard Discrete Diffusion baselines. Furthermore, our approach exhibits superior training efficiency, achieving higher success rates with fewer training iterations, particularly in challenging long-horizon tasks.
	\end{itemize}

	\section{Related Work}
	\label{sec:headings}
	\subsection{Diffusion Language Models \& Block Diffusion}
	
	Discrete Diffusion Language Models~\citep{dlm:d3pm,dlm:diffusionbert,dlm:dream,dlm:llada,dlm:llada2,dlm:ssd,dlm:sedd} have emerged as a promising new paradigm, aiming to support parallel decoding through bidirectional sequence modeling. This approach fundamentally mitigates the compounding execution errors of Autoregressive (AR) generation and significantly accelerates inference. However, the actual speedup of diffusion models is inherently bottlenecked by the iterative denoising process, which requires a large Number of Function Evaluations (NFE). Furthermore, the bidirectional attention mechanism sacrifices the causal structure, precluding the reuse of past information via KV-cache. To resolve these inefficiencies, Block Diffusion~\citep{bd:block,dlm:fastdllm,bd:fastdlm2} was introduced. By maintaining an autoregressive structure at the block level while performing parallel diffusion within each block, it successfully preserves both causality and KV-cache compatibility, ultimately achieving much faster and more practical parallel decoding. In this work, we extend Block Diffusion to VLA models to optimize the learning of multi-modal action sequences.
	
	\subsection{Adapting Autoregressive Models to Diffusion}
	
	Despite the theoretical advantages of diffusion, the vast majority of prevailing Large Language and Vision-Language Models~\citep{llm:llama,llm:qwen,vlm:llava,vlm:qwen3,vlm:internvl,vlm:paligemma,vlm:prismatic,vlm:openflamingo} are pretrained using the autoregressive paradigm, leaving a severe scarcity of strong diffusion-native backbones. This is largely because training Diffusion Language Models (dLLMs) from scratch is significantly more challenging than the standard AR paradigm; current state-of-the-art dLLMs trained from scratch often lag behind their AR counterparts in both data scale and model parameters. To harness parallel acceleration without discarding the immense world knowledge already embedded in AR models, adapting pretrained AR models~\citep{ar2dlm:scaling} into diffusion models has become a vital new paradigm. Nevertheless, bridging the gap between the causal next-token prediction of AR and the bidirectional denoising objective of diffusion is notoriously difficult. Recent studies indicate that Block Diffusion serves as a natural interpolation between these two paradigms. Compared to direct diffusion fine-tuning , adapting an AR model through a block-wise semi-autoregressive~\citep{dlm:llada2,ar2dlm:ar2drunway,ar2dlm:efficient,ar2dlm:sdar,ar2dlm:nbdiff} approach ensures a much smoother transition towards dLLMs while better preserving pretrained capabilities. Distinct from these language-centric approaches, we extend this adaptation paradigm to the robotics VLA domain, pioneering the transition of large-scale pretrained AR VLAs into efficient Discrete Diffusion policies via Block Diffusion.
    
	\subsection{Discrete Vision-Language-Action (VLA) Models}

     Inspired by LLM advances, VLA models have emerged as a promising paradigm for general-purpose robotic manipulation~\citep{vla:cogact,vla:pi05,cui2025openhelix,bai2026reshaping,bai2026latent}. A major line of work focuses on discrete VLA modeling~\citep{vla:openvla,wang2025vqvla,vla:spatialvla,vla:worldvla,zhao2025vlas}, where AR backbones generate tokenized robot actions. While systems like OpenVLA demonstrate strong transfer capabilities, their token-by-token decoding introduces high latency and compounding errors. Recent studies have explored diffusion-based VLA generation~\citep{ddvla:fastdvla,ddvla:dream,ddvla:lladavla,ddvla:udvla} to address these issues, yet many directly adopt dLLM formulations that treat perception and action as a homogeneous sequence. Although DDVLA~\citep{ddvla:discrete} adapts AR-based VLAs into discrete diffusion, its reliance on extensive iterative denoising limits efficiency. In this work, we leverage the asymmetric nature of perception and action by treating visual-linguistic features as stable prefix anchors. By tailoring a block-wise causal structure with a Diffusion Forcing objective, we ensure that parallel action denoising remains strictly grounded in the multi-modal context. This design enables continuous prefix KV-cache reuse, achieving a hardware-friendly balance between reasoning depth and execution reactivity.

	\section{Method}
	\label{sec:method}
    
\begin{figure}[t]
		\centering
		\includegraphics[width=0.99\linewidth]{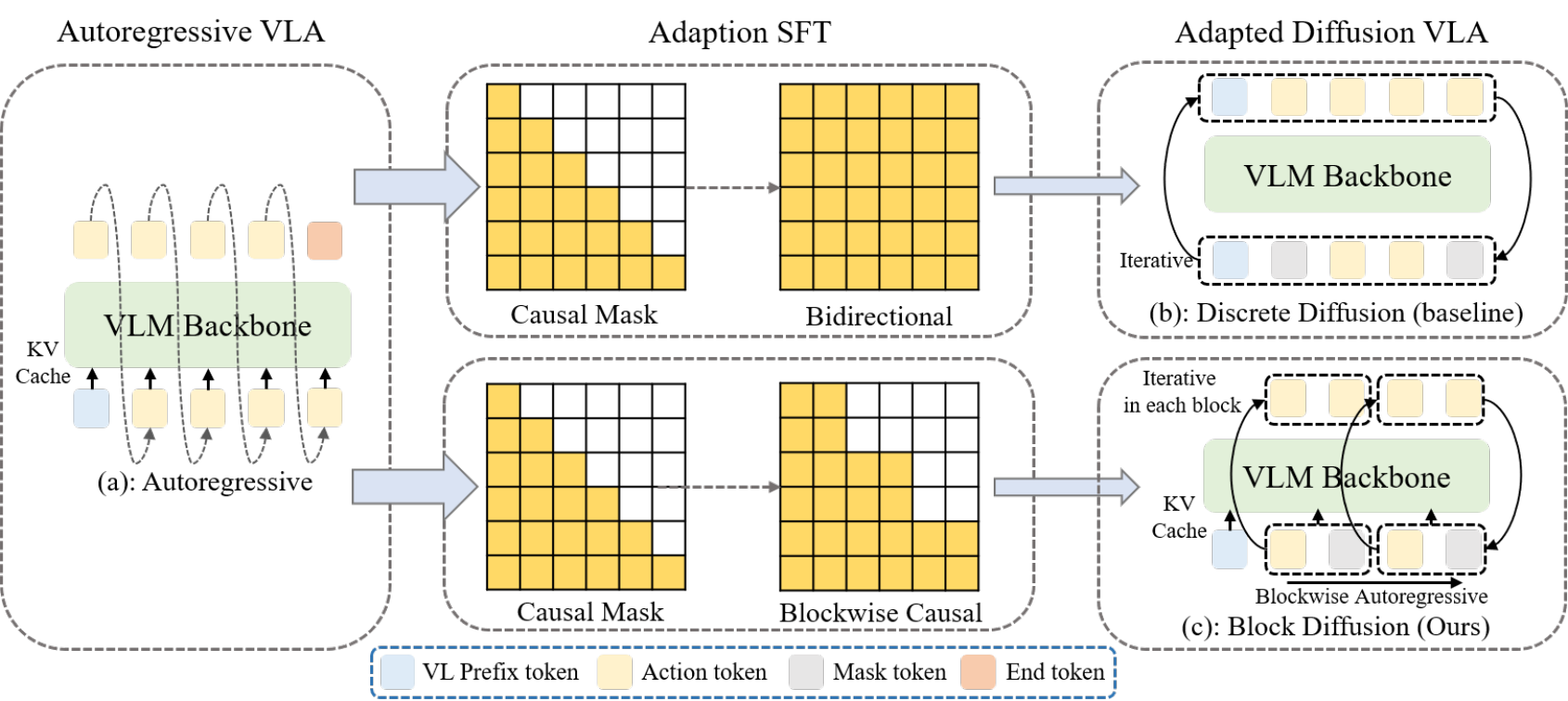}
		\caption{Overall architecture of the proposed BlockVLA framework. The adaptation process is illustrated in three stages. \textbf{Left}: We initialize our model from a pretrained Autoregressive VLA, inheriting its robust vision-language alignment and causal reasoning capabilities. \textbf{Middle}: During the Adaption SFT phase, we illustrate the shift in training objectives. While the baseline transitions from a standard causal mask to a fully bidirectional mask (top), our method employs a Blockwise causal masking strategy (bottom). This encourages intra-block parallel denoising while preserving global causal dependencies. \textbf{Right}: Adapted Diffusion VLA during inference. In contrast to standard Discrete Diffusion, which requires full-sequence iterations, our Block Diffusion maintains an autoregressive flow at the block level, enabling efficient KV Cache reuse and significantly reducing the deployment latency of discrete diffusion-based robotic policies.}
		\label{fig:overall_framework}
	\end{figure}

    As illustrated in~\Cref{fig:overall_framework}, BlockVLA is a semi-autoregressive discrete diffusion framework for robotic action generation. 
    It preserves autoregressive dependencies across action blocks while enabling parallel denoising within each block, thereby combining global temporal consistency with efficient local refinement. 
    We first review the preliminaries of discrete diffusion and block diffusion language modeling in~\Cref{subsec:preliminaries}, and then formulate their extension to VLA policy learning. 
    We then present the architecture of BlockVLA in~\Cref{subsec:arch}, including the pretrained backbone, action tokenization, block-wise diffusion formulation, masking strategy, and token-shift design. 
    Finally,~\Cref{subsec:pipeline} describes the block-wise training and inference pipeline.

	\subsection{Preliminaries: Discrete and Block Diffusion Language Model}\label{subsec:preliminaries}
	Unlike continuous diffusion models~\citep{dm:ddpm,dm:ddim} that operate in continuous space using Gaussian noise $\mathbf x_t = \alpha_t \mathbf x_0 + \sigma_t \mathbf \epsilon$ where $\mathbf \epsilon \sim \mathcal N(0, \mathbf I)$ , Discrete Diffusion Language Models~\citep{dlm:d3pm,dlm:llada,dlm:llada2,dlm:ssd} (dLLMs) define the diffusion process directly on the discrete vocabulary $\mathcal{V}$ of size $V$. Let $\mathbf{x}_0 = [\mathbf{x}_0^{(1)}, \dots, \mathbf{x}_0^{(L)}]$ denote a sequence of $L$ tokens, where each $\mathbf{x}_0^{(i)} \in \{0, 1\}^{1 \times V}$ is a one-hot row vector. The forward process independently corrupts each token, where the probability of the intermediate state $\mathbf{x}_t = [\mathbf{x}_t^{(1)}, \dots, \mathbf{x}_t^{(L)}]\in \mathbb R^{L\times V}$ is given by:
	
	\begin{equation}
        q(\mathbf{x}_t \mid \mathbf{x}_0) = \prod_{i=1}^L q(\mathbf x_t^{(i)} \mid \mathbf x_0^{(i)}) =\prod_{i=1}^L \mathrm{Categorical}\left(\mathbf{x}_t^{(i)}; p = \mathbf{x}_0^{(i)} \bar{\mathbf{Q}}_t\right), \quad \text{with} \quad \bar{\mathbf{Q}}_t = \bar{\alpha}_t \mathbf{I} + (1-\bar{\alpha}_t)\mathbf{1}\mathbf{m}^{\top},
    \end{equation}
	
	where $\bar{\alpha}_t$ denotes the retention coefficient determined by the noise schedule, $\mathbf{1}\in \mathbb R^{V\times 1}$ denotes the all-one vector, and $\mathbf{m}\in \mathbb R^{V\times 1}$ denotes the one-hot vector corresponding to the absorbing \textbf{[MASK]} token. Under this formulation, $\bar{\mathbf{Q}}_t \in \mathbb{R}^{V \times V}$ ensures that each token either stays in its original state or transitions to the mask state with probabilities defined by the schedule. The model $p_\theta(\mathbf x_0 \mid \mathbf x_t)$ is then trained to predict the original tokens at the masked positions. The training objective minimizes the cross-entropy loss over masked tokens:
	
    \begin{equation}
        \mathcal{L}(\theta)
        =
        -\mathbb{E}_{t,\mathbf{x}_0,\mathbf{x}_t}
        \left[
        \frac{1}{t}
        \sum_{i=1}^{L}
        \mathbf{1}\!\left[\mathbf{x}_t^{(i)}=\textbf{[MASK]}\right]
        \log p_\theta(\mathbf{x}_0^{(i)} \mid \mathbf{x}_t)
        \right].
    \end{equation}
	
	where $\mathbf{1}[\cdot]$ is the indicator function that isolates the loss to the masked indices, and the factor $1/t$ follows the standard reweighting used in masked dLLMs. During inference, the generation process departs from the standard token-by-token autoregressive suite. It begins with a sequence entirely composed of \textbf{[MASK]} tokens. In each iteration, the model performs a parallel forward pass to predict probability distributions for all masked positions. Following a predefined noise schedule, only a subset of tokens with the highest prediction confidence is retained and fixed, while remaining uncertain positions are re-masked for subsequent refinement. This iterative "predict-and-refine" strategy allows the model to utilize bidirectional context and perform robust error correction, ultimately leading to a coherent and precise action sequence within a few steps.

	\textbf{Block Diffusion}~\citep{bd:block} emerges as a semi-autoregressive paradigm combining the advantages of global coherence and parallel generation efficiency. In this paradigm, the clean sequence $\mathbf{x}_0$ is partitioned into $B$ ordered blocks $[\mathbf{x}_0^1, \mathbf{x}_0^2, \dots, \mathbf{x}_0^B]$, where each block $\mathbf{x}_0^b$ contains $L'$ tokens $\mathbf{x}_0^b = [(\mathbf{x}_0^b)^{(1)}, (\mathbf{x}_0^b)^{(2)}, \dots, (\mathbf{x}_0^b)^{(L')}]$. The generation follows a hybrid scheme: \textbf{autoregressive modeling between blocks and discrete diffusion within each block}. The joint likelihood of the sequence factorizes as:
	
    \begin{equation}
        \log p_\theta(\mathbf{x}_0)
        =
        \sum_{b=1}^{B}
        \log p_\theta(\mathbf{x}_0^b \mid \mathbf{x}_0^{<b}).
    \end{equation}
	
	For each block $\mathbf{x}_0^b$, the conditional distribution is modeled using a discrete diffusion process, where we independently assign timesteps $(t_1,t_2,\dots,t_B)$ to the $B$ blocks. We distinguish two conditioning strategies for the block-level history:
	\begin{itemize}[leftmargin=1.5em]
		\item \textbf{Teacher Forcing:}~\citep{bd:block,ar2dlm:sdar} The model conditions on the clean ground-truth previous blocks, which is formulated as $p_\theta(\mathbf{x}_0^b \mid \mathbf{x}_{t_b}^b, \mathbf{x}_0^{<b})$. This ensures stable training by providing an uncorrupted history. 
		\item \textbf{Diffusion Forcing:}~\citep{bd:df,bd:diffusionforce,bd:fastdlm2} The model conditions on potentially ``noisy'' previous blocks from earlier diffusion steps, formulated as $p_\theta(\mathbf{x}_0^b \mid \mathbf{x}_{t_b}^b, \mathbf{x}_{t'}^{<b})$. This aligns the training distribution with the iterative refinement during inference, enhancing the model's robustness to past prediction errors.
	\end{itemize}
	
	For a sequence of length $L$ partitioned into $B$ blocks, each block with length $L'=L/B$, assuming $L$ is divisible by $B$.  The overall learning objective $\mathcal{L}_{\text{BD}}$ is defined as the average masked denoising loss across all blocks:
	
    \begin{equation}
        \mathcal{L}_{\text{BD}}(\theta)
        =
        \frac{1}{B}
        \sum_{b=1}^{B}
        \mathbb{E}_{t_b,\mathbf{x}_0,\mathbf{x}_{t_b}^b}
        \left[
        -\frac{1}{t_b}
        \sum_{i=1}^{L'}
        \mathbf{1}\!\left[(\mathbf{x}_{t_b}^b)^{(i)}=\textbf{[MASK]}\right]
        \log p_\theta\!\left(
        (\mathbf{x}_0^b)^{(i)}
        \mid
        \mathbf{x}_{t_b}^b,\mathbf{h}^{<b}
        \right)
        \right],
    \end{equation}
    	
    where $\mathbf{1}[\cdot]$ isolates the loss to the masked indices within block $b$, and $\mathbf{h}^{<b}$ denotes the block-level history: $\mathbf{h}^{<b}=\mathbf{x}_0^{<b}$ for Teacher Forcing and $\mathbf{h}^{<b}=\mathbf{x}_{t'}^{<b}$ for Diffusion Forcing. This unified architecture breaks the autoregressive bottleneck by allowing parallel token generation within blocks while maintaining the structural consistency of the global sequence.

\subsection{Architecture of BlockVLA}\label{subsec:arch}

    \subsubsection{Backbone and Action Tokenization}
    BlockVLA builds upon Discrete Diffusion VLA (DDVLA)~\citep{ddvla:discrete} and the multimodal architecture of OpenVLA~\citep{vla:openvla}. As illustrated in~\Cref{fig:overall_framework}, we preserve the Prismatic-7B~\citep{vlm:prismatic} backbone, utilizing SigLIP~\citep{vlm:siglip} and DINO v2~\citep{vlm:dinov2} to extract complementary visual representations that are linearly projected into the Llama~\citep{llm:llama} embedding space. While OpenVLA generates actions autoregressively and DDVLA uses full-sequence discrete diffusion, BlockVLA retains the unified transformer structure but introduces a block-wise formulation. By enforcing block-wise causal constraints over action tokens, our framework enables efficient diffusion-based action generation.
    
    Following the action discretization scheme used in RT-2~\citep{vla:rt2} and OpenVLA~\citep{vla:openvla}, we discretize each continuous action dimension into 256 bins using quantile-based binning. The bin boundaries are computed from the 1st to 99th percentiles of the action distribution to reduce the influence of outliers. The binary gripper command is represented as an independent token. Each single-timestep action is represented by 7 tokens, including 3 translation tokens, 3 rotation tokens, and 1 gripper token. Given an action horizon $H$ and action dimension $D$, we flatten the action chunk into a one-dimensional action sequence $\mathbf{a}$ of length $H D$ and concatenate it with the multimodal prefix:
    \begin{equation}
    \mathbf{x} =
    [\textbf{BOS}, \mathbf{v}, \mathbf{p}, \mathbf{l},
    a_1^{(1)}, \dots, a_1^{(D)}, \dots,
    a_H^{(1)}, \dots, a_H^{(D)}, \textbf{EOS}],
    \end{equation}
    where $\mathbf{v}$, $\mathbf{p}$, and $\mathbf{l}$ denote the visual, proprioceptive, and language tokens, respectively. The token $a_h^{(d)}$ denotes the discretized value of the $d$-th action dimension at timestep $h$. We use $\mathbf{c}=[\textbf{BOS},\mathbf{v},\mathbf{p},\mathbf{l}]$ to denote the multimodal prefix context, and reserve $\mathbf{a}$ for the flattened action token sequence.
    
	\begin{figure}[t]
		\centering
		\captionsetup[subfigure]{labelformat=simple, labelsep=space}
		\renewcommand\thesubfigure{(\arabic{subfigure})}
		
		\begin{subfigure}{0.48\textwidth}
			\centering
			\includegraphics[width=\linewidth]{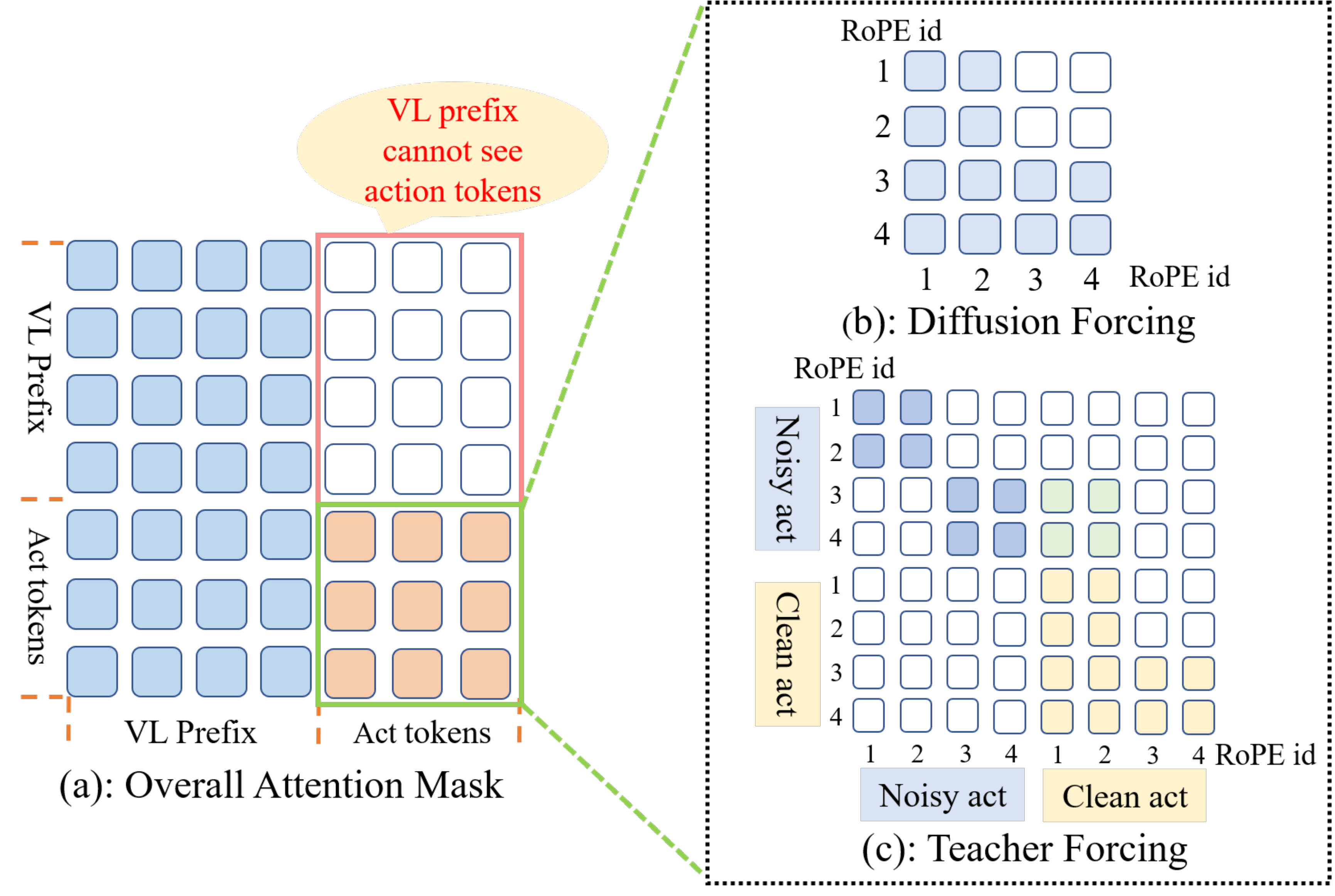}
			\caption{Block-wise masking strategies for Teacher Forcing and Diffusion Forcing.}
			\label{fig:mask_strategy} 
		\end{subfigure}
		\hfill
		\begin{subfigure}{0.48\textwidth}
			\centering
			\includegraphics[width=\linewidth]{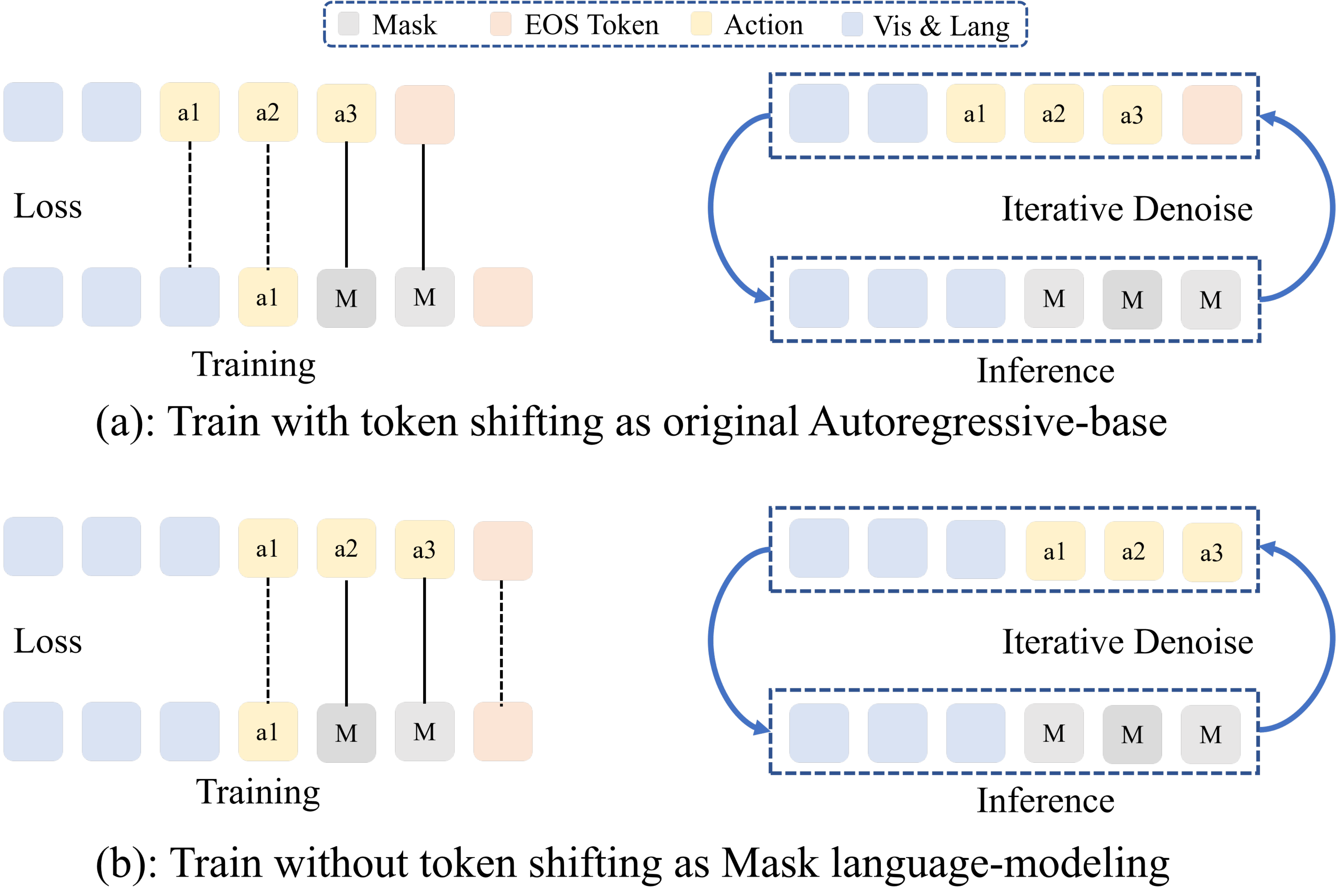}
			\caption{Training and inference with or without token shift.}
			\label{fig:token_shift} 
		\end{subfigure}
		
		\caption{Visualization of the block-wise masking strategy and token-shift design in BlockVLA.}
		\label{fig:mask_shift_comparison}
	\end{figure}

    \subsubsection{Block Diffusion Formulation of BlockVLA}
    To improve the efficiency of diffusion-based action generation, BlockVLA partitions the flattened action sequence $\mathbf{a}$ into $B$ consecutive blocks, denoted as $\mathbf{a} = [\mathbf{a}^1,\mathbf{a}^2,\ldots,\mathbf{a}^B]$. Let $\mathbf{a}^{b}$ denote the $b$-th action block, and let $\mathbf{a}^{<b}$ denote all preceding action blocks. The model maintains an autoregressive dependency across blocks while performing parallel denoising within each block. This design differs from standard full-sequence discrete diffusion, where the entire action sequence is repeatedly refined. By restricting iterative denoising to local blocks, BlockVLA reduces the effective computation associated with repeated denoising steps and allows previously generated blocks to serve as fixed causal context.
    
    During generation, completed blocks are treated as prefix context for subsequent blocks. Therefore, the hidden states associated with the multimodal prefix and completed action blocks can be reused through prefix KV caching. In contrast, conventional bidirectional diffusion decoding updates the entire sequence during each denoising step, making direct reuse of standard KV-cache states difficult. BlockVLA thus provides a practical compromise between autoregressive decoding and diffusion-based parallel refinement: global temporal coherence is preserved through block-level causality, while local action tokens are generated in parallel within each block, effectively balancing computational efficiency with high-fidelity trajectory prediction.
	
	\subsubsection{Masking Strategy}
    We implement BlockVLA using a structured block-wise attention mask. As shown in~\Cref{fig:mask_strategy}, we consider two training variants that differ in the context exposed to the current noisy action block: Teacher Forcing in (c) conditions on clean preceding blocks, whereas Diffusion Forcing in (b) conditions on noisy or partially denoised preceding blocks.

    In the \textbf{Teacher Forcing} variant~\citep{bd:block}, the $b$-th block is trained to predict the clean target block from its noisy version, conditioned on the prefix context and \textbf{clean history blocks}:
    \begin{equation}
    p_{\theta}(\mathbf{a}_0^b \mid \mathbf{a}_{t_b}^b, \mathbf{a}_0^{<b}, \mathbf{c}),
    \end{equation}
    where $\mathbf{a}_{t_b}^b$ is the noisy current block at noise level $t_b$, $\mathbf{a}_0^b$ is its clean target, $\mathbf{a}_0^{<b}$ denotes clean preceding action blocks, and $\mathbf{c}=[\textbf{BOS},\mathbf{v},\mathbf{p},\mathbf{l}]$ is the multimodal prefix. To expose the clean history without leaking the current or future targets, we concatenate the noisy action sequence and the clean action sequence:
    \begin{equation}
    \mathbf{x}_{\mathrm{input}} =
    [\textbf{BOS}, \mathbf{v}, \mathbf{p}, \mathbf{l},
    \mathbf{a}_{\mathbf{t}}, \textbf{EOS}, \mathbf{a}_0],
    \end{equation}
    where $\mathbf{a}_{\mathbf{t}}=[\mathbf{a}_{t_1}^1,\ldots,\mathbf{a}_{t_B}^B]$ and $\mathbf{t}=(t_1,\ldots,t_B)$ indicate that each block can be corrupted with an independent noise level, and $\mathbf{a}_0$ denotes the clean action sequence. The attention mask allows tokens in $\mathbf{a}_{t_b}^b$ to attend to $\mathbf{c}$ and $\mathbf{a}_0^{<b}$, but prevents access to $\mathbf{a}_0^{\geq b}$. Meanwhile, the Teacher Forcing implementation requires two additional details to ensure consistency between training and inference:
    \begin{itemize}[leftmargin=1.5em]
        \item \textbf{Loss masking.} The appended clean sequence $\mathbf{a}_0$ is used only as auxiliary conditioning context. It does not participate in the diffusion loss, and its CE targets are set to the ignore index $-100$.
        \item \textbf{Position-index alignment.} The appended clean sequence should not be treated as a new temporal continuation of the noisy sequence. Therefore, corresponding tokens in $\mathbf{a}_{\mathbf{t}}$ and $\mathbf{a}_0$ are assigned identical RoPE position indices~\citep{llm:rope}. Equivalently, the position ids of the clean suffix are shifted back to match the action positions in $\mathbf{a}_{\mathbf{t}}$, so that clean previous blocks provide KV context at the correct action positions.
    \end{itemize}

    In the \textbf{Diffusion Forcing} variant~\citep{bd:df}, no clean action suffix is appended. Thus, we only need to corrupt each block in the action sequence.
    \begin{equation}
    \mathbf{x}_{\mathrm{input}} =
    [\textbf{BOS}, \mathbf{v}, \mathbf{p}, \mathbf{l},
    \mathbf{a}_{\mathbf{t}}, \textbf{EOS}].
    \end{equation}
    The $b$-th block is instead trained as
    \begin{equation}
    p_{\theta}(\mathbf{a}_0^b \mid \mathbf{a}_{t_b}^b, \mathbf{a}_{t'}^{<b}, \mathbf{c}),
    \end{equation}
    where $\mathbf{a}_{t'}^{<b}$ denotes preceding blocks that may have different noise levels and may remain noisy or partially denoised. Thus, Diffusion Forcing differs from Teacher Forcing in the source of block-level history: it conditions on imperfect previous blocks rather than clean previous blocks. This better matches the inference-time pattern, where previously generated blocks may contain residual prediction errors.

    Finally, we impose a \textbf{constraint on the prefix context} $\mathbf{c}$. Tokens in $\mathbf{c}$ use bidirectional attention to support multimodal fusion among \textbf{BOS}, visual, proprioceptive, and language tokens. However, prefix tokens are not allowed to attend to action tokens. This constraint prevents leakage from future actions, keeping the conditioning pattern consistent between training and inference: all action blocks condition on the full multimodal prefix, but the prefix representation is computed independently of the action sequence. Compared with the bidirectional mask in DDVLA, our mask preserves parallel denoising within each block while introducing causal dependencies across action blocks.

	\subsubsection{Token Shift}
    Standard autoregressive language models are trained with a one token shift as shown in ~\Cref{fig:token_shift}, where the hidden state at position $n-1$ predicts the token at position $n$. Prior work on adapting pretrained AR models to diffusion language modeling~\citep{ar2dlm:scaling} suggests that preserving this shift may help align diffusion training with pretrained AR representations. DDVLA~\citep{ddvla:discrete} follows this design.

    
	\begin{wrapfigure}{r}{0.50\textwidth}
		\centering
		\vspace{-13pt} 
		\captionof{table}{Analysis of token shift on \textit{LIBERO-Object} for DDVLA baseline (Success Rate \%).}
		\label{tab:token_shift_ablation}
		\begin{small}
			\begin{tabular}{@{}ccccccc@{}}
				\toprule
				\multirow{2}{*}{\textbf{Variant}} & \multicolumn{6}{c}{\textbf{Training Steps}} \\
				\cmidrule(lr){2-7}
				& \textbf{5k} & \textbf{10k} & \textbf{15k} & \textbf{20k}  & \textbf{30k} & \textbf{40k}  \\ \midrule
				w token shift & 51.2 & 56.6 & 80.2 & 90.6 & 95.0  & 94.2  \\
				\textbf{w/o token shift} & \textbf{59.6} & \textbf{66.0} & \textbf{81.0} & \textbf{94.2} & \textbf{95.2} & \textbf{95.8} \\ \bottomrule
			\end{tabular}
		\end{small}
		\vspace{-14pt} 
	\end{wrapfigure}
    
    We empirically revisit this choice for VLA policy learning. As shown in~\Cref{tab:token_shift_ablation}, which aligns with recent findings in Efficient-DLM~\citep{ar2dlm:efficient}, removing the token shift does not degrade performance on \textit{LIBERO-Object}; instead, it yields higher success rates throughout most training stages. This result suggests that after diffusion fine-tuning, the model can recover action tokens at their original positions without relying on the AR-style shifted prediction target. Therefore, unless otherwise specified, we remove the token-shift operation in both our DDVLA baseline and BlockVLA, which directly align each prediction target with its corresponding input position. 
	
	\subsection{Training and Inference Pipeline of BlockVLA}\label{subsec:pipeline}
	
	To support efficient training and low-latency robotic control, BlockVLA adopts a block-wise generation pipeline for discrete diffusion VLA. The key idea is to decompose an action chunk into a sequence of blocks, where temporal consistency is modeled through autoregressive dependencies across blocks, and parallel refinement is performed within each block via discrete diffusion. The overall procedure is summarized in Algorithm~\ref{algo:block_diffusion_all}.

    \subsubsection{Training Phase}
    During training, we partition the tokenized action sequence $\mathbf{a}$ into $B$ consecutive blocks, denoted as $[\mathbf{a}^1,\ldots,\mathbf{a}^B]$. For each block $\mathbf{a}^b$, we independently sample a diffusion timestep $t_b \sim \mathcal{U}(0,1)$ and apply the forward masking process to obtain its noised version $\mathbf{a}_{t_b}^b$. Within each block, bidirectional attention is used to model dependencies among action dimensions and timesteps. Across blocks, we impose a block-wise causal mask such that current block prediction can only attend to the multimodal prefix and preceding blocks. In our default Diffusion Forcing setting, preceding blocks are provided as noised blocks $\mathbf{a}_{t'}^{<b}$, leading to the training objective
    \begin{equation}
        p_{\theta}(\mathbf{a}_0^b \mid \mathbf{a}_{t_b}^b, \mathbf{a}_{t'}^{<b}, \mathbf{c}),
    \end{equation}
    where $\mathbf{c}$ denotes the multimodal prefix consisting of \textbf{BOS}, visual, proprioceptive, and language tokens. This block-wise mask allows all blocks to be trained in parallel while preserving the inference-time causal structure. Since the full action sequence is available during training, we do not use KV caching, which avoids serializing block computation and enables efficient batched gradient updates.
    
    \subsubsection{Inference Phase}
    During inference, BlockVLA generates action blocks sequentially while denoising tokens within each block in parallel. We first encode the multimodal prefix $\mathbf{c}$ and store its key-value states in the KV cache. For the $b$-th block, we initialize a block of \textbf{[MASK]} tokens and perform iterative discrete denoising conditioned on the cached prefix and previously generated blocks. During the refinement of the current block, the cached states remain fixed, since only the current block tokens are updated. After the block is fully denoised, its final tokens are forwarded once to compute the corresponding key-value states, which are then appended to the cache and used as causal context for the next block. This procedure enables prefix KV-cache reuse across completed blocks, reducing redundant computation compared with full-sequence bidirectional diffusion decoding. The positional encoding indices are updated after each completed block so that subsequent blocks maintain the correct temporal ordering.
	
	\begin{minipage}[t]{0.49\textwidth}
		\begin{algorithm}[H]
			
			\caption{Block Diffusion Training Process}
			\label{algo:block_diffusion_all}
			
			\begin{algorithmic}[1]
				\REQUIRE blocks $B$, noise process $q_t(\cdot\mid\mathbf{a})$, model $\mathbf{x}_\theta$, loss $\mathcal{L}_{\text{BD}}$, VL prefix $\mathbf{c}=[\textbf{BOS}, \mathbf{v},\mathbf{p},\mathbf{l}]$, action  $\mathbf{a}_0$
				\REPEAT
				\STATE Sample $t_1, \dots, t_B \sim U[0, 1]$
				\FOR{$b=1$ \TO $B$}
				\STATE $\mathbf{a}_{t_b}^b \sim q_{t_b}(\cdot\mid\mathbf{a}^b)$ \algcomment{Add noise to each block}
				\ENDFOR
				\STATE $\mathbf{a}_{\mathbf{t}} \leftarrow [\mathbf{a}_{t_1}^1,\dots,\mathbf{a}_{t_B}^B]$
				\STATE $\mathbf{x}_{\text{input}} \leftarrow [\mathbf{c}, \mathbf{a}_{\mathbf{t}},\textbf{EOS}, \mathbf{a}_0]$ if \textbf{Teacher Forcing} else $[\mathbf{c}, \mathbf{a}_{\mathbf{t}}, \textbf{EOS}]$
				\STATE  \algcomment{Forward with asymmetric block-wise mask}
				\STATE $\mathbf{a}_{\text{logits}} \leftarrow \mathbf{x}_\theta(\mathbf{x}_{\text{input}})$
				\STATE Take gradient step on $\nabla_\theta \mathcal{L}_{\text{BD}}(\mathbf{a}_{\text{logits}}, \mathbf{a}_0; \theta)$
				\UNTIL{converged}
			\end{algorithmic}
		\end{algorithm}
	\end{minipage}
	\hfill
	\begin{minipage}[t]{0.50\textwidth}
		\begin{algorithm}[H]
			\caption{Block Diffusion Inference Process}
			\begin{algorithmic}[1]
				\REQUIRE blocks $B$, model $\mathbf{x}_\theta$, sampling algorithm \text{SAMPLE}, VL prefix $\mathbf{c}=[\textbf{BOS}, \mathbf{v},\mathbf{p},\mathbf{l}]$, KV-Cache $\mathbf K, \mathbf V$
				\STATE $\mathbf{a} \leftarrow \emptyset$
				\STATE  \algcomment{Pre-fill KV cache with prefix}
				\STATE $(\cdot), \mathbf K, \mathbf V \leftarrow \mathbf{x}_\theta(\mathbf{c}, \text{update=True})$ 
				\FOR{$b=1$ \TO $B$}
				\STATE  \algcomment{Denoise action block $b$ using fixed $\mathbf K, \mathbf V$}
				\STATE $\mathbf{a}^b, (\cdot), (\cdot) \leftarrow \text{SAMPLE}(\mathbf{x}_\theta, \mathbf K, \mathbf V, \text{update=False})$
				\STATE  \algcomment{Update cache once the action block is fixed}
				\STATE $(\cdot),\mathbf K, \mathbf V \leftarrow \mathbf{x}_\theta(\mathbf{a}^b, \text{update=True})$
				\STATE $\mathbf{a} \leftarrow \mathbf{a} \oplus \mathbf{a}^b$
				\STATE Update position embedding indices
				\ENDFOR
				\RETURN $\mathbf{a}$
			\end{algorithmic}
		\end{algorithm}
	\end{minipage}
	
	\section{Experiments}
	\subsection{Experiment Setup}

    \subsubsection{Benchmarks}
    We evaluate our method on two widely-used robotic simulation benchmarks: LIBERO~\citep{bench:libero} and SimplerEnv~\citep{bench:simplerenv}.
     LIBERO comprises four specialized task suites, including \textit{Spatial, Object, Goal}, and \textit{Long}, designed to assess policy generalization across diverse spatial layouts, object variations, semantic goals, and extended horizons. Each suite encompasses 10 distinct tasks, with 50 episodic demonstrations provided per task (totaling 500 episodes per suite) for evaluation. We report \textbf{Success Rate} for the LIBERO benchmark. SimplerEnv is a high-fidelity real-to-sim benchmark that provides a simulated counterpart for physical robotic setups. We conduct evaluations on the \textbf{WidowX} platform across four pick-and-place tasks. Following the standard protocol, we evaluate each task over 24 episodes and report two key metrics: \textbf{Grasp Count} and \textbf{Success Count}. 
     
	\subsubsection{Baselines}
	We select \textbf{OpenVLA}~\citep{vla:openvla} and \textbf{Discrete Diffusion VLA(DDVLA)}~\citep{ddvla:discrete} as our primary baselines. OpenVLA serves as the representative autoregressive baseline, utilizing a Prismatic-7B backbone that integrates SigLIP and DINOv2 as dual visual encoders with a Llama-based language model to generate action tokens. DDVLA inherits this multimodal architecture but reformulates action generation as a discrete diffusion process. Following our preliminary findings in~\Cref{tab:token_shift_ablation}, which suggest that token shifting provides marginal gains for diffusion-based VLA, we adopt the \textbf{no token shift} configuration for all DDVLA and BlockVLA results to ensure a robust and fair comparison.


	\subsection{Main Results}
	  Our main experimental results across the LIBERO benchmarks and the SimplerEnv WidowX robot tasks are detailed in ~\Cref{tab:results_comparison} and ~\Cref{tab:simpler_env}, respectively. \textbf{BlockVLA} demonstrates a clear performance advantage over both the autoregressive and discrete diffusion baselines. In the LIBERO suites, our method achieves an average success rate of \textbf{91.7\%} within 50k steps, consistently outperforming OpenVLA and showing significantly faster convergence than the standard DDVLA. In the SimplerEnv real-to-sim evaluations, BlockVLA maintains competitive performance across diverse pick-and-place scenarios. Notably, while significantly reducing the number of denoising steps compared to the diffusion baseline, our framework still achieves comparable or slightly superior success rates, validating its reliability in more complex environmental settings. Specifically, our BlockVLA exhibits significant advantages in two primary aspects:

    \begin{table*}[t]
		\centering
		\begin{minipage}[c]{0.67\linewidth}
			\centering
			\caption{Success Rate (\%) across training steps on LIBERO benchmarks. We compare our 2-step per block denoising against the 12-step Discrete Diffusion VLA (DDVLA)~\citep{ddvla:discrete} baseline and Autoregressive OpenVLA~\citep{vla:openvla}. The results of OpenVLA~\citep{vla:openvla} is quoted from the original paper.}
			\label{tab:results_comparison}
			\begin{small}
				\resizebox{\linewidth}{!}{%
					\begin{tabular}{@{}llccccccc@{}}
						\toprule
						\multirow{2}{*}{\textbf{Method}} & \multirow{2}{*}{\textbf{Type}} & \multicolumn{7}{c}{\textbf{Success Rate (\%) at Training Steps}} \\
						\cmidrule(lr){3-9}
						&  & 5k & 10k & 15k & 20k & 30k & 40k & 50k \\
						\midrule
						\rowcolor[gray]{0.9} \textit{LIBERO-Spatial} & & & & & & & & \\
						\rule{0pt}{2ex}OpenVLA~\citep{vla:openvla} & Autoregressive & \multicolumn{7}{c}{\cellcolor{blue!10} 84.7 (Reported by~\citep{vla:openvla})} \\
						DDVLA~\citep{ddvla:discrete} & Discrete Diffusion & 48.0 & 56.8 & 68.4 & 81.0 & 88.2 & 87.2 & 89.8 \\
						\textbf{BlockVLA (Ours)} & \textbf{Block Diffusion} & \textbf{53.8} & \textbf{79.2} & \textbf{81.6} & \textbf{83.0} & \textbf{88.4} & \textbf{89.0} & \textbf{90.6} \\
						\midrule
						\rowcolor[gray]{0.9} \textit{LIBERO-Object} & & & & & & & & \\
						\rule{0pt}{2ex}OpenVLA~\citep{vla:openvla} & Autoregressive & \multicolumn{7}{c}{\cellcolor{blue!10} 88.4 (Reported by~\citep{vla:openvla})} \\
						DDVLA~\citep{ddvla:discrete} & Discrete Diffusion & 59.6 & 66.0 & 81.0 & 94.2 & 95.2 & 95.8 & 96.6 \\
						\textbf{BlockVLA (Ours)} & \textbf{Block Diffusion} & \textbf{69.2} & \textbf{76.6} & \textbf{94.0} & \textbf{94.8} & \textbf{97.0} & \textbf{96.8} & \textbf{97.6} \\
						\midrule
						\rowcolor[gray]{0.9} \textit{LIBERO-Goal} & & & & & & & & \\
						\rule{0pt}{2ex}OpenVLA~\citep{vla:openvla} & Autoregressive & \multicolumn{7}{c}{\cellcolor{blue!10} 79.2 (Reported by~\citep{vla:openvla})} \\
						DDVLA~\citep{ddvla:discrete} & Discrete Diffusion & 51.2 & 69.2 & 77.6 & 89.6 & 91.8 & \textbf{94.2} & 92.6 \\
						\textbf{BlockVLA (Ours)} & \textbf{Block Diffusion} & \textbf{60.0} & \textbf{75.6} & \textbf{86.0} & \textbf{89.8} & \textbf{92.6} & 93.2 & \textbf{93.2} \\
						\midrule
						\rowcolor[gray]{0.9} \textit{LIBERO-Long} & & & & & & & & \\
						\rule{0pt}{2ex}OpenVLA~\citep{vla:openvla} & Autoregressive & \multicolumn{7}{c}{\cellcolor{blue!10} 53.7 (Reported by~\citep{vla:openvla})} \\
						DDVLA~\citep{ddvla:discrete} & Discrete Diffusion & 0.4 & 3.6 & 3.8 & 18.6 & 23.2 & 37.2 & 53.6 \\
						\textbf{BlockVLA (Ours)} & \textbf{Block Diffusion} & \textbf{35.8} & \textbf{44.2} & \textbf{62.2} & \textbf{70.2} & \textbf{69.6} & \textbf{78.6} & \textbf{85.2} \\
						\midrule
						\rowcolor[gray]{0.8} \textbf{Average} & & & & & & & & \\
						\rule{0pt}{2ex}OpenVLA~\citep{vla:openvla} & Autoregressive & \multicolumn{7}{c}{\cellcolor{blue!10} 76.5 (Reported by~\citep{vla:openvla})} \\
						DDVLA~\citep{ddvla:discrete} & Discrete Diffusion & 39.8 & 48.9 & 57.7 & 70.9 & 74.6 & 78.6 & 83.2 \\
						\textbf{BlockVLA (Ours)} & \textbf{Block Diffusion} & \textbf{54.7} & \textbf{68.9} & \textbf{81.0} & \textbf{84.5} & \textbf{87.0} & \textbf{89.4} & \textbf{91.7} \\
						\bottomrule
					\end{tabular}%
				}
			\end{small}
		\end{minipage}
		\hfill
		\begin{minipage}[c]{0.29\linewidth}
			\centering
			\includegraphics[width=\linewidth]{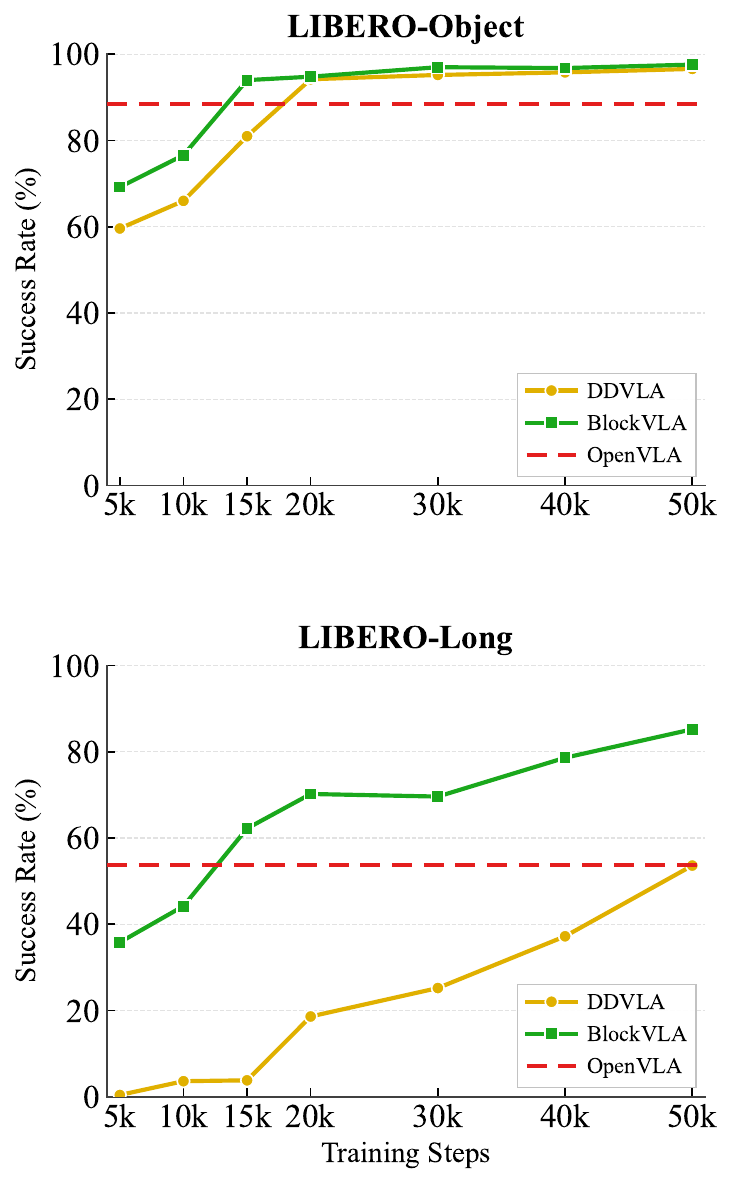}
			\captionof{figure}{Success Rate (\%) vs. training steps on \textit{LIBERO-Object} and \textit{LIBERO-Long}. Our BlockVLA reaches high success rate faster than baseline.}
			\label{fig:libero_curve}
		\end{minipage}
	\end{table*}

\begin{table}[htbp]
	\centering
	\caption{SimplerEnv WidowX robot results (partial) over 4 pick-and-place tasks. Success Count (out of 24 episodes) is used as the metric. We train for 60k steps and save checkpoints every 10k steps.}
	\label{tab:simpler_env}
	\small
    \vskip 0.05in
	\resizebox{\columnwidth}{!}{%
		\begin{tabular}{@{}llcccccccc@{}}
			\toprule
			\multirow{2}{*}{\textbf{Task}} & \multirow{2}{*}{\textbf{Method}} & \multicolumn{2}{c}{\textbf{10k}} & \multicolumn{2}{c}{\textbf{20k}} & \multicolumn{2}{c}{\textbf{40k}} & \multicolumn{2}{c}{\textbf{60k}} \\
			\cmidrule(lr){3-4} \cmidrule(lr){5-6} \cmidrule(lr){7-8} \cmidrule(lr){9-10}
			&  & Grasp & Success & Grasp & Success & Grasp & Success & Grasp & Success \\
			\midrule
			\multirow{2}{*}{Put Carrot on Plate}
			& DDVLA~\citep{ddvla:discrete}          & 10/24 & 3/24  & \textbf{13/24} & \textbf{6/24}  & 13/24 & 3/24 & 17/24 & 5/24 \\
			& \textbf{BlockVLA (Ours)} & \textbf{15/24} & \textbf{6/24} & 12/24 & 6/24 & 12/24 & 2/24 & 20/24 & 4/24 \\
			\midrule
			\multirow{2}{*}{Put Spoon on Towel}
			& DDVLA~\citep{ddvla:discrete}          & 12/24 & 3/24  & \textbf{17/24}& \textbf{7/24}  & 14/24 & 10/24 & 13/24 & 7/24 \\
			& \textbf{BlockVLA (Ours)} & 17/24 & 5/24 & 15/24 & 9/24 & 17/24 & 9/24 & \textbf{17/24} & \textbf{12/24} \\
			\midrule
			\multirow{2}{*}{Stack Green on Yellow}
			& DDVLA~\citep{ddvla:discrete}          & 11/24 & 1/24  & \textbf{13/24} & \textbf{4/24}  &  15/24 & 1/24 & 12/24 & 3/24 \\
			& \textbf{BlockVLA (Ours)} & \textbf{15/24} & \textbf{4/24} & 13/24 & 4/24 & 18/24 & 3/24 & 14/24 & 3/24 \\
			\midrule
			\multirow{2}{*}{Put Eggplant in Basket}
			& DDVLA~\citep{ddvla:discrete}          & 17/24 & 8/24  & 19/24 & 12/24 & 19/24 & 14/24 & \textbf{21/24} & \textbf{15/24} \\
			& \textbf{BlockVLA (Ours)} & \textbf{22/24} & \textbf{15/24} & 16/24 & 10/24 & 18/24 & 14/24 & 20/24 & 15/24 \\
			\bottomrule
		\end{tabular}%
	}
\end{table}

	\subsubsection{Enhanced Training Efficiency} 
	
\begin{wrapfigure}{r}{0.45\textwidth}
		\centering
		\vspace{-1pt}
		\includegraphics[width=0.43\textwidth]{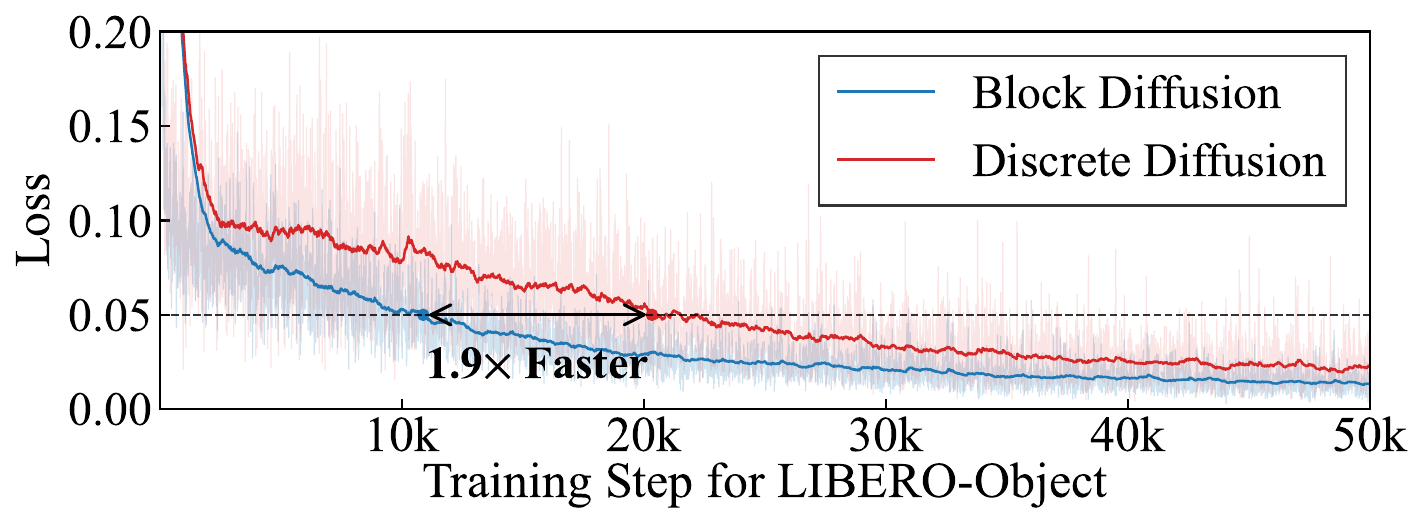}
		\caption{Training dynamics on \textit{LIBERO-Object}. Our method achieves faster convergence and more stable loss reduction compared to the baseline.}
		\label{fig:loss_curve}
		\vspace{-1pt}
	\end{wrapfigure}

	As illustrated in~\Cref{fig:loss_curve} (taking \textit{LIBERO-Object} as a representative), Block Diffusion exhibits significantly superior training dynamics compared to the discrete diffusion baseline. When evaluating the Action Loss with EMA smoothing (thresholded at 0.05), our framework achieves faster convergence and maintains a more stable, lower-variance loss trajectory. This enhanced efficiency facilitates rapid adaptation during fine-tuning, which is particularly critical for long-horizon tasks. 
	
	For instance, in the challenging \textit{LIBERO-Long} shown in~\Cref{fig:libero_curve} while the baseline struggles to bootstrap (merely 3.8\% success rate at 15k steps), our BlockVLA effectively captures temporal dependencies to reach \textbf{60.0\%} within the same interval, demonstrating its high-efficiency adaptation. The similar trend of training efficiency is also observed on SimplerEnv, where BlockVLA consistently outperforms the DDVLA baseline at the early 10k-step checkpoint across all four evaluation tasks (see~\Cref{tab:simpler_env}).

	\subsubsection{Inference Acceleration} 
	Block Diffusion delivers a transformative leap in deployment efficiency by synergizing parallel decoding with KV-cache reuse. As shown in ~\Cref{fig:overall_results}(b), our framework, utilizing only 2 denoising steps per block, achieves a remarkable inference throughput of \textbf{186.7 tokens/s}. This represents a \textbf{3.3$\times$} speedup over the Discrete Diffusion baseline and an \textbf{8.0$\times$} improvement relative to the Autoregressive OpenVLA. These results validate that Block Diffusion effectively mitigates the computational bottleneck inherent in iterative diffusion processes, making high-performance VLA models viable for high-frequency, real-time robotic control.

	\begin{figure}[htbp]
		\centering
		\begin{minipage}{0.31\textwidth}
			\centering
			\includegraphics[width=\linewidth]{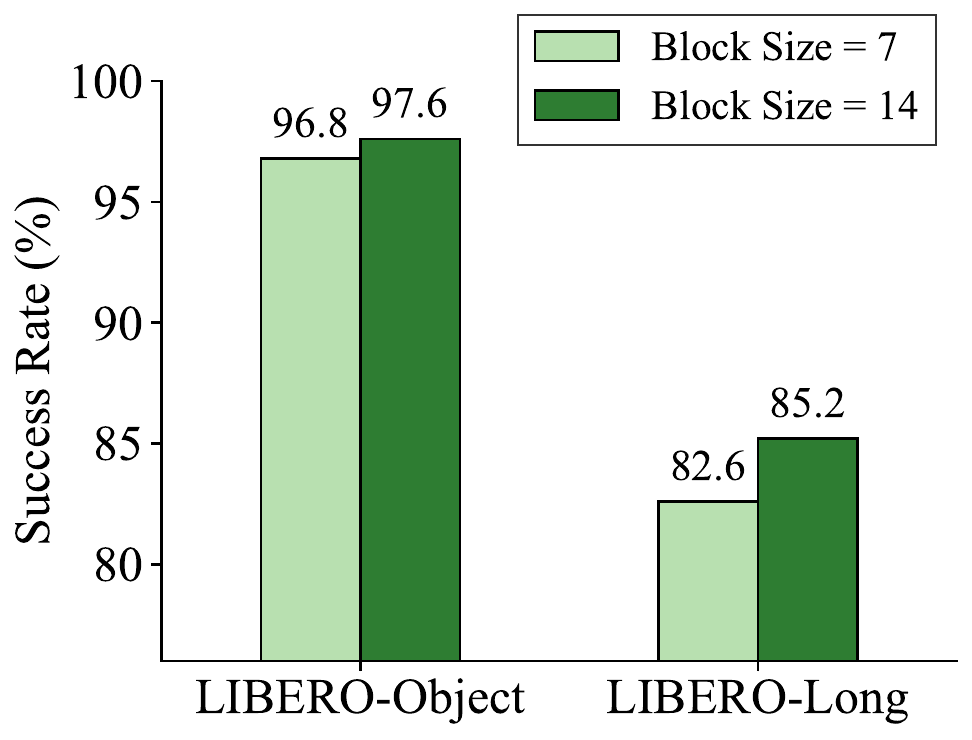}
			\caption{Ablation study on block size at 50k training steps for \textit{LIBERO-Object} and \textit{LIBERO-Long}. We observe that block size 14 yields better performance.}
			\label{fig:block_size}
		\end{minipage}
		\hfill 
		\begin{minipage}{0.31\textwidth}
			\centering
			\includegraphics[width=\linewidth]{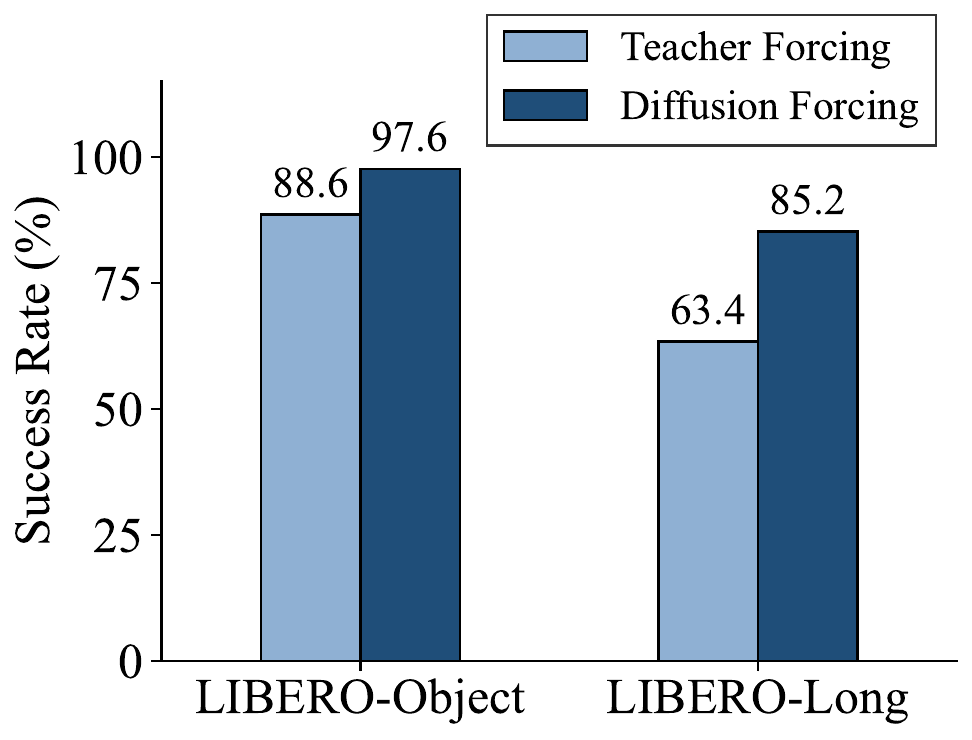}
			\caption{Comparison of Teacher Forcing and Diffusion Forcing at training step 50k, demonstrating the stability of our Diffusion Forcing training strategy.}
			\label{fig:forcing_comparison}
		\end{minipage}
        \hfill 
        \begin{minipage}{0.30\textwidth}
            \centering
			\includegraphics[width=\linewidth]{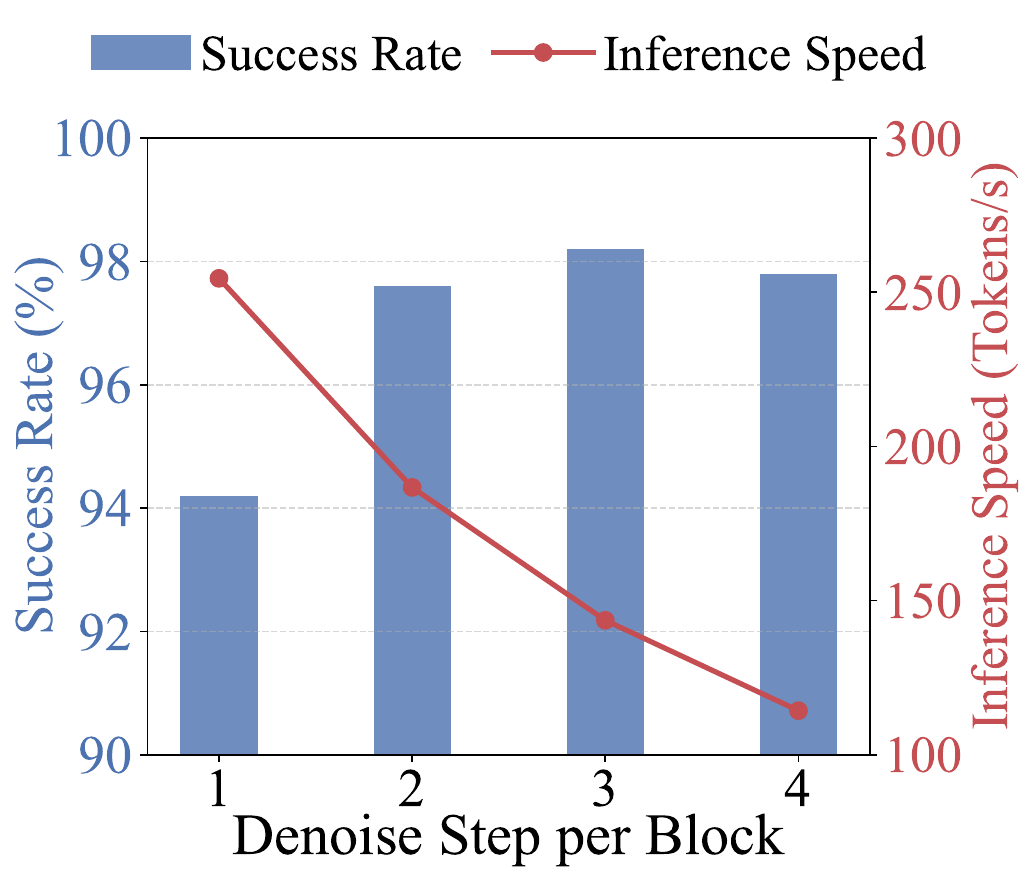}
			\caption{Ablation study on denoise steps per block under block at 50k training steps for \textit{LIBERO-Object}. We observe a performance-speed trade-off as iterations increase.}
			\label{fig:denoise_step_analysis}
        \end{minipage}
	\end{figure}
	
	\subsection{Further Analysis \& Ablation Study}
	
	To investigate the inner workings of BlockVLA, we perform a systematic analysis on \textit{LIBERO-Object} and \textit{LIBERO-Long}. We conduct an ablation study on key architectural components as follows.

  \subsubsection{Impact of Block Size}
  We investigate the sensitivity of our framework to the block size $B$ by evaluating performance on \textit{LIBERO-Object} and \textit{LIBERO-Long} with $B=7$ and $B=14$. As illustrated in ~\Cref{fig:block_size}, $B=14$ consistently outperforms $B=7$ across both benchmarks. This performance gap suggests that while $B=7$ corresponds to a single-step action, the larger block size $B=14$ enables the model to jointly model the distribution of two consecutive time steps within a single denoising process. By capturing these inter-step dynamic correlations and temporal dependencies, the parallel diffusion head ensures better motion continuity and trajectory smoothness, which are essential for maintaining stability in complex, long-horizon manipulation tasks.
	
	\subsubsection{Diffusion Forcing vs. Teacher Forcing}
	Following the masking strategies illustrated in~\Cref{fig:mask_strategy} (b)(c), we compare the performance of \textbf{Teacher Forcing} and \textbf{Diffusion Forcing} across \textit{LIBERO-Object} and \textit{LIBERO-Long}. As previously shown in~\Cref{fig:forcing_comparison} , \textbf{Diffusion Forcing} consistently outperforms the standard \textbf{Teacher Forcing} baseline. This suggests that in robotic VLA tasks, modeling the $b$-th block based on corrupted (noisy) previous states facilitates a more robust coupling between causal blocks compared to training on ground-truth (clean) history. This mechanism effectively bridges the gap between training-time supervision and inference-time autoregressive error accumulation.
	
	
	\subsubsection{Impact of Denoising Steps per Block}
	We further analyze the influence of the number of denoising iterations within each block using the \textit{LIBERO-Object} as shown in~\Cref{fig:denoise_step_analysis}. With a fixed block size of 14, we evaluate the performance as the denoising steps vary from 1 to 4. As depicted in the accompanying figure, the success rate exhibits a notable peak at 3 denoising steps. This result reveals a critical \textbf{performance-efficiency trade-off}: while increasing the number of iterations enhances the fidelity of action generation, it simultaneously incurs higher computational latency, thereby reducing the inference throughput. In our framework, 2 or 3 steps provide an optimal balance, achieving high-success rate while maintaining real-time inference speeds.

	\section{Conclusion}
	
	In this paper, we presented \textbf{BlockVLA}, a block diffusion framework that bridges the gap between causal reasoning and parallel action generation in robotic policies. By restructuring the action sequence into a hierarchical block-wise architecture, our method successfully reconciles the temporal consistency and KV-Cache efficiency of autoregressive models with the high-throughput advantages of discrete diffusion. Our extensive evaluations on the LIBERO and SimplerEnv benchmarks demonstrate that Block Diffusion achieves superior \textbf{inference efficiency}, delivering a $3.3\times$ speedup over standard diffusion baselines while maintaining high execution precision. Furthermore, we highlight a significant gain in \textbf{training efficiency}; our framework facilitates a smoother adaptation from pretrained Vision-Language Models (VLMs), converging substantially faster and requiring less data than bidirectional diffusion approaches. These results underscore the potential of Block Diffusion as a scalable and efficient paradigm for deploying large-scale pretrained models into real-world robotic applications. Future work will explore the integration of this block-based diffusion strategy with multi-modal sensor fusion to further enhance the robustness of long-horizon robotic manipulation.

\clearpage
\bibliographystyle{iclr2025_conference}
\bibliography{main}

\end{document}